\documentclass[letterpaper,10pt,conference]{ieeeconf}
\IEEEoverridecommandlockouts
\usepackage[T1]{fontenc}
\usepackage{graphicx}
\usepackage{capt-of}
\usepackage{amsmath,amssymb}
\usepackage{booktabs}
\usepackage{xcolor}
\usepackage{array,colortbl}
\usepackage{tikz}
\usetikzlibrary{arrows.meta,positioning,calc}
\usepackage{url}
\usepackage{balance}
\DeclareRobustCommand{\name}{\mbox{\ensuremath{\phi}-RIE}}
\definecolor{scene}{HTML}{245C7D}
\definecolor{evidence}{HTML}{97612E}
\definecolor{physics}{HTML}{3D7352}
\definecolor{appearance}{HTML}{70558E}

\usepackage{flushend}
\newcounter{phirieSavedDblTopNumber}
\AddToHook{shipout/after}[phirie-front-floats]{%
  \ifnum\value{page}=2\relax
    \setcounter{dbltopnumber}{\value{phirieSavedDblTopNumber}}%
    \RemoveFromHook{shipout/after}[phirie-front-floats]%
  \fi
}
\title{\LARGE\bfseries\boldmath \name: From \underline{Ph}otoreal\underline{i}stic \underline{R}econstruction to \underline{I}nteractive \underline{E}nvironments}
\author{%
Runyi Yang$^{1}$, Deheng Zhang$^{1}$, Xiaoye Wang$^{1}$,
Kanzhi Wu$^{2}$, Lei Sun$^{1}$,\\
Ajad Chhatkuli$^{1}$, Kunyu Peng$^{3,*}$,
Luc Van Gool$^{1}$, Danda Pani Paudel$^{1}$ \\
\textbf{Project Page}: \href{https://insait-institute.github.io/PhiRIE/}{https://github.com/insait-institute/PhiRIE} 
\thanks{$^{1}$ INSAIT, Sofia University ``St. Kliment Ohridski''.}%
\thanks{$^{2}$ vivo Mobile Communication Co., Ltd.}%
\thanks{$^{3}$ Karlsruhe Institute of Technology, $^{*}$Corresponding author.}%
}

\usepackage[pagebackref=false, breaklinks=true, colorlinks, bookmarks=false]{
        hyperref
}
\definecolor{hanpurple}{HTML}{5218FA}
\hypersetup{
        colorlinks=true,
        linkcolor={blue},
        citecolor={hanpurple},
        urlcolor={magenta}
}

\IEEEaftertitletext{
\begin{minipage}{\textwidth}
\vspace{-1em}
\centering
\includegraphics[width=0.95\textwidth]{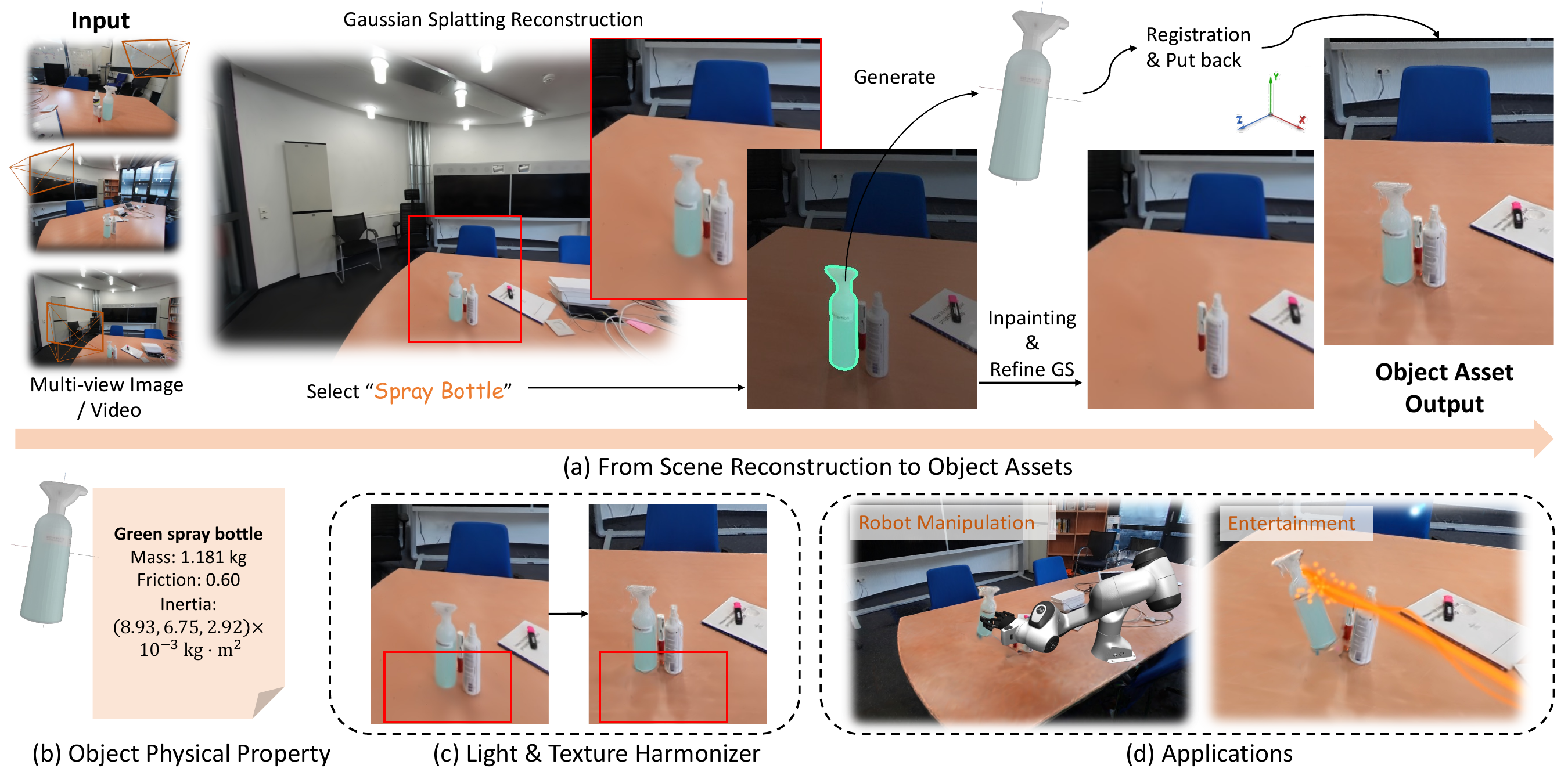}
\vskip-2ex
\captionof{figure}{\textbf{From captured appearance to physical interaction.}
\name\ converts captured Gaussians into an Interactive Environment through
Scene Observation and Coupled Scene Construction. Shared object identities
connect completed assets to source-Gaussian removal and background completion.
Panel~(c) motivates optional appearance harmonization after rendering to
address lighting and shadow mismatch following object insertion or motion.}
\label{fig:teaser}
\end{minipage}
\par\vspace{0.5\baselineskip}
}
\begin{document}
\maketitle
\thispagestyle{empty}
\pagestyle{empty}
\providecommand{\phirieYes}{\textcolor{green!40!black}{\ensuremath{\checkmark}}}
\providecommand{\phirieNo}{\textcolor{red!65!black}{\ensuremath{\times}}}
\providecommand{\phirieUnknown}{\textcolor{black!55}{---}}
\begin{table*}[!t]
\centering\footnotesize
\setlength{\tabcolsep}{2.0pt}
\renewcommand{\arraystretch}{1.13}
\caption{\textbf{Scene-conversion and rendering capabilities.} Checks indicate included operations, crosses absent operations, and dashes unspecified evidence.}
\vskip-2ex
\label{tab:system-capabilities}
\begin{tabular*}{\textwidth}{@{\extracolsep{\fill}}l*{7}{c}>{\columncolor{violet!6}}c@{}}
\toprule
Property & \shortstack{SplatSim\cite{splatsim}} & \shortstack{Re$^3$Sim\cite{re3sim}} &
\shortstack{HoloScene\cite{holoscene}} & \shortstack{PolaRiS\cite{polaris}} &
\shortstack{SimRecon\cite{simrecon}} & \shortstack{GASE\cite{gase}} &
\shortstack{SimFoundry\cite{simfoundry}} & \shortstack{\textbf{\name}\textbf{(ours)}} \\
\midrule
Rigid-body simulation & \phirieYes & \phirieYes & \phirieYes & \phirieYes & \phirieYes & \phirieYes & \phirieYes & \phirieYes \\
Gaussian background & \phirieYes & \phirieYes & \phirieYes & \phirieYes & \phirieNo & \phirieYes & \phirieYes & \phirieYes \\
Gaussian movable objects & \phirieYes & \phirieNo & \phirieYes & \phirieUnknown & \phirieNo & \phirieUnknown & \phirieNo & \phirieYes \\
Generative object completion & \phirieNo & \phirieNo & \phirieYes & \phirieYes & \phirieYes & \phirieYes & \phirieYes & \phirieYes \\
Exposed-background synthesis & \phirieNo & \phirieNo & \phirieUnknown & \phirieNo & \phirieUnknown & \phirieYes & \phirieYes & \phirieYes \\
Evidence-based 3D selection & \phirieNo & \phirieNo & \phirieYes & \phirieNo & \phirieUnknown & \phirieNo & \phirieNo & \phirieYes \\
\midrule
\textbf{GS full rendering mode} & \phirieYes & \phirieNo & \phirieYes & \phirieUnknown & \phirieNo & \phirieUnknown & \phirieNo & \phirieYes \\
\bottomrule
\end{tabular*}
\vskip-4ex
\end{table*}

\begin{figure*}[t]
\centering
\includegraphics[width=\textwidth]{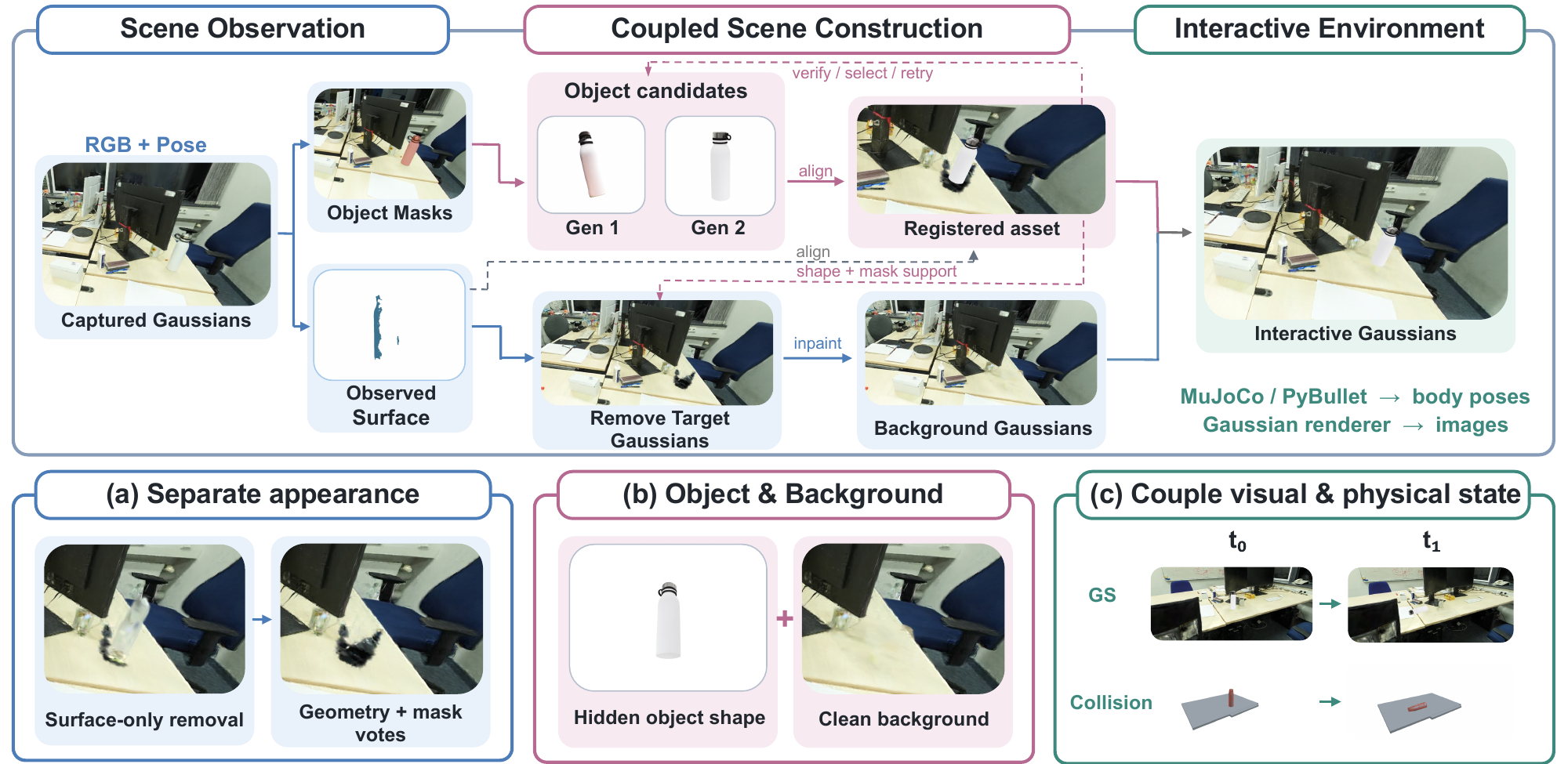}
\vskip-2ex
\caption{\textbf{Overview of \name.} Scene Observation provides object masks and observed surfaces. Coupled Scene Construction produces registered assets and background Gaussians. The Interactive Environment renders their composition from simulator body poses. The lower panels illustrate (a) appearance separation, (b) object and background completion, and (c) coupled visual and physical state.}
\vskip-4ex
\label{fig:system}
\end{figure*}

\raggedbottom
\begin{abstract}
3D Gaussian Splatting (3DGS) can reconstruct a captured scene photorealistically, but the resulting representation does not by itself support physical interaction. Robot simulation instead requires object-level change, \textit{i.e.}, objects must move independently, make contact, and reveal previously occluded surroundings. This gap arises because object appearance may remain entangled with the background, while hidden object geometry and occluded background content may be unobserved.
To address this challenge, we present \name, a Gaussian-native pipeline that converts selected objects into movable simulator assets while preserving the remaining reconstruction. Our key observation is that asset construction and source removal should be coupled, \textit{i.e.}, one object identity should define the movable asset and the scene content to remove and complete. Accordingly, Scene Observation supplies shared evidence to Coupled Scene Construction, which creates registered assets and completed background Gaussians for simulator-driven rendering in an Interactive Environment. This coupling preserves unedited Gaussians while aligning visual and physical state.
On 50 ScanNet++ scenes, evidence-based selection and registration retry increase matched F1 at 20\,mm from 0.336 to 0.383 at fixed retention. Further tests demonstrate asset executability, manipulation gains over a single-generator baseline, and the visual cost of conversion. Together, these results demonstrate that \name\ enables interactive scene conversion.
\end{abstract}

\vspace{-1em}
\section{Introduction}
\label{sec:intro}
Robot simulation requires executable scene content, \textit{i.e.}, objects with identities, metric placements, collision geometry, and appearance that follows their motion. Benchmarks such as LIBERO and BEHAVIOR provide this structure through prepared assets~\cite{libero,omnigibson}. In contrast, real-world capture can preserve the layout, clutter, and appearance of an actual environment. Among current scene representations, 3D Gaussian Splatting (3DGS) is particularly attractive for photorealistic real-to-sim because it combines high-fidelity reconstruction with efficient rendering and has already been integrated into robot simulators~\cite{kerbl2023,splatsim,re3sim,discoverse}. However, converting such a reconstruction into an interactive simulation environment requires more than photorealistic rendering.

The underlying difficulty is that 3DGS primitives are optimized to explain images rather than represent independently movable physical objects. Consider a robot lifting a cup from a table. To support this interaction, the visual representation of the cup must be separated from its surroundings and move with its physical body. Moreover, its hidden shape must support contact, and the previously covered tabletop must become visible after the cup moves. Yet an accurate rendering of the initial scene does not guarantee these properties. The cup and table can be explained by overlapping Gaussians, while the base of the cup and the tabletop beneath it may remain unobserved. This example reveals three coupled requirements for conversion. First, \emph{separation} assigns captured appearance to independently movable objects. Second, \emph{completion} supplies missing object surfaces for contact and background content exposed by motion. Finally, \emph{state consistency} binds visual and collision representations to a common coordinate frame and drives them with the same simulated motion. These requirements must be addressed jointly, because inserting an asset without removing its original appearance creates a duplicate, while erasing it without completing the background leaves holes.

To address these coupled requirements, we introduce \name\ (Fig.~\ref{fig:teaser}), which organizes scene conversion into the three stages shown in Fig.~\ref{fig:system}. \emph{Scene Observation} extracts object masks and an observed surface from the captured views. \emph{Coupled Scene Construction} then uses this shared evidence to address separation and completion. Its object branch generates, aligns, verifies, and selects completed asset candidates, with registration retry for uncertain orientation. Meanwhile, its background branch uses the same object's shape and masks to remove source Gaussians and complete the exposed region. Consequently, each selected asset corresponds to the appearance removed from the scene, while unedited Gaussians retain their captured appearance.

The \emph{Interactive Environment} addresses state consistency by combining the constructed assets and background with simulator state. Specifically, simulator body poses drive the corresponding Gaussian assets, while completed meshes provide collision geometry without replacing Gaussian appearance. As a result, rendered objects remain aligned with their simulated bodies (Fig.~\ref{fig:system}(c)). After rendering, an optional harmonization stage reduces the residual lighting and shadow mismatch illustrated in the teaser (Fig.~\ref{fig:teaser}(c), Sec.~\ref{sec:harmonization}) without changing the constructed scene or its dynamics.


To assess whether the resulting scene representations are both faithful and executable, we evaluate \name across successive stages of conversion. We evaluate construction on 50 ScanNet++ scenes by measuring candidate retention and geometric accuracy against independent references, and separately quantify the visual cost of conversion on held-out views~\cite{scannetpp}. We then test physical validity through simulator export and isolated-body drop tests, and interaction utility through controlled asset-replacement experiments in RoboCasa. By evaluating these stages separately, we avoid conflating construction availability with reconstruction fidelity, physical validity, or downstream task success.

Our contributions are summarized:
\begin{itemize}
    \item A Gaussian-native pipeline that preserves unedited captured
    appearance while turning selected objects into independently
    movable assets.

    \item Coupled Scene Construction, in which shared object evidence
    connects candidate alignment and selection to source-Gaussian
    removal and background completion.

    \item A real-scene evaluation spanning construction availability,
    geometric and visual fidelity, physical validity, and
    manipulation performance.
\end{itemize}
\section{Related Work}
\label{sec:related}
\textbf{Gaussian Splatting for robotics.}
Neural mapping recovers geometry for perception and planning~\cite{imap,niceslam,isdf}, while 3DGS and 2DGS emphasize captured appearance and efficient rendering~\cite{kerbl2023,huang2024}.
PhysGaussian models dynamics with Gaussian primitives, and Robo-GS combines Gaussian appearance, meshes, and physical attributes~\cite{physgaussian,robogs}.
SplatSim, DISCOVERSE, and GSWorld integrate Gaussian rendering with physics simulation~\cite{splatsim,discoverse,jiang2025gsworld}.
\name{} focuses on converting a captured Gaussian scene into independently movable objects and a completed background.

\textbf{Real-to-sim scene construction and evaluation.}
Re$^3$Sim and SimFoundry pair Gaussian backgrounds with mesh-rendered objects, while HoloScene binds Gaussian appearance to completed meshes~\cite{re3sim,simfoundry,holoscene}.
SimRecon constructs compositional assets from video, GASE reconstructs foreground and background after image-space separation and completion, and SimuScene refines generated shapes and layouts from a single image using physics feedback~\cite{simrecon,gase,lee2026simuscene}.
RialTo learns policies in reconstructed environments, while DexNinja uses simulation for contact-rich policy learning~\cite{rialto,lou2026dexninja}.
LIBERO, BEHAVIOR, and RoboCasa provide prepared tasks and interactive assets~\cite{libero,omnigibson,robocasa365}, while SIMPLER and PolaRiS emphasize matched observations and control for policy evaluation~\cite{simpler,polaris}.
In contrast, \emph{Coupled Scene Construction} edits an existing Gaussian field locally, using shared object evidence for candidate alignment and selection, source removal, and background completion while preserving unrelated primitives.
Table~\ref{tab:system-capabilities} compares construction and rendering choices.
Its \emph{GS full rendering mode} uses Gaussians for both the background and movable objects.

\textbf{Object generation and scene editing.}
TRELLIS, TRELLIS.2, ReconViaGen, and SAM 3D Objects generate assets from images~\cite{trellis,trellis2,reconviagen,sam3dobjects}, but do not jointly address metric placement, collision preparation, and removal of captured appearance.
SAM3 provides concept-conditioned masks, while Chorus encodes semantic and instance cues in Gaussian scenes~\cite{sam3,chorus}.
GaussianEditor and image inpainting edit appearance~\cite{gaussianeditor,lama}, whereas DiffusionHarmonizer enhances renderings~\cite{harmonizer}.
In \name{}, background completion updates scene Gaussians, while optional harmonization changes only rendered images.

\section{Gaussian-Native Scene Conversion}
\label{sec:method}

\textbf{Overview.}
As shown in Fig.~\ref{fig:system}, \name\ converts a captured Gaussian scene
into movable object assets and a completed background through
\emph{Scene Observation}, \emph{Coupled Scene Construction}, and an
\emph{Interactive Environment}. Shared object instances connect asset
generation and registration with source-Gaussian removal and background
completion. We assemble the assets for physics simulation and drive their
Gaussian appearance with the simulated body poses. Optional appearance
harmonization follows rendering and addresses the distinct appearance issue
in Fig.~\ref{fig:teaser}(c).

\subsection{Scene Representation}
\label{sec:representation}
Inputs are a Gaussian reconstruction $G^0$, calibrated images
$\mathcal V=\{I_k,K_k,T_k\}$ with camera intrinsics $K_k$ and camera-to-world poses
$T_k$ in a common metric frame, and an aligned scene surface from a scan or fused depths rendered
from $G^0$. For editable object $i$, $Q_i$ contains observed surface
vertex samples, $G_i$ is its visual Gaussian asset in a canonical object frame,
and $S_i(t)$ is its object-to-scene transformation at timestamp $t$, with
$S_i(0)\in\mathrm{Sim}(3)$.
Removing the instance's source Gaussians $R_i\subseteq G^0$ and adding background
completion $G^{\rm fill}$ gives
\begin{equation}
\begin{split}
G^{\rm bg}&=\left(G^0\setminus\bigcup_i R_i\right)\cup G^{\rm fill},\\
G(t)&=G^{\rm bg}\cup\bigcup_i\mathcal W(S_i(t),G_i).
\end{split}
\label{eq:scene}
\end{equation}
Here, $\mathcal W(S,G)$ transforms Gaussian set $G$ by $S$.
The background retains source Gaussians outside the removal sets,
while object assets move independently. These correspond to the
Background $G^{\rm bg}$ and Interactive
$G(t)$ in Fig.~\ref{fig:system}.

\subsection{Scene Observation}
\label{sec:discovery}
From the captured Gaussians and calibrated RGB + Pose inputs,
Scene Observation supplies the Object Masks and Observed Surface.
Frozen SAM3 produces image masks~\cite{sam3}, which we lift onto the
scene surface using first-hit ray intersections. We associate these
regions across views by voxel overlap to obtain object instances with
surface samples $Q_i$, masks, and spatial bounds.

\subsection{Coupled Scene Construction}
\label{sec:coupled}
The two branches share object identity but complete different missing
content: object shape and the exposed background (Fig.~\ref{fig:system}(b)).
A registered asset also supplies shape support for source-Gaussian removal.

\subsubsection{Object candidates and registered assets}
\label{sec:construction}
\textbf{Generate and align candidates.}
The default pool combines single-view TRELLIS~\cite{trellis} and multi-view
ReconViaGen~\cite{reconviagen}, each providing a completed mesh
and visual Gaussians. Generation predicts missing shape, while
observations constrain metric placement.
For candidate surface samples $P_i$, we estimate scale, rotation, and
translation $S$. We initialize scale from robust observed dimensions,
search yaw under an upright hypothesis, and refine with
partial-to-complete iterative closest point (ICP), followed by scale
and translation refinement.
We score alignment with a symmetric clipped nearest-point distance:
\begin{equation}
E_i(S)=d_{\tau}(SP_i,Q_i)+d_{\tau}(Q_i,SP_i),
\label{eq:registration}
\end{equation}
where
\begin{equation}
d_{\tau}(A,B)=\frac{1}{|A|}\sum_{a\in A}\min\!\left(\tau,\min_{b\in B}\|a-b\|_2\right).
\label{eq:clipped}
\end{equation}
The two directions penalize unsupported candidate surfaces and
unexplained observations. Clipping at $\tau$ limits outliers and
penalties on unobserved regions. This measures observed agreement,
not hidden-shape correctness.

\textbf{Verify, select, and retry.}
Candidates are ranked using construction checks for registration,
scale, observation support, collision validity, and isolated settling,
followed by residual and stability criteria
(Sec.~\ref{sec:implementation}).
If initial candidates fail, a bounded retry tests alternative source-up
directions. It changes the alignment hypothesis without regenerating
or repairing shape.
The default constructor keeps the best numerically valid candidate
and records unresolved checks. A strict variant rejects candidates
that still fail, separating the effects of registration retry from rejection.

\subsubsection{Background Gaussians}
\label{sec:background}
\textbf{Remove source appearance.}
The background branch uses the same identity as the registered asset.
Figure~\ref{fig:system}(a) contrasts surface-only removal with geometry
and mask votes. Surface proximity can miss diffuse or low-opacity object contributions,
so we combine three sets of source Gaussians:
\begin{equation}
R_i=R_i^{\rm obs}\cup R_i^{\rm asset}\cup R_i^{\rm mask}.
\label{eq:removal}
\end{equation}
$R_i^{\rm obs}$ and $R_i^{\rm asset}$ select Gaussians near the observed
surface and registered completed asset, respectively.
The latter extends coverage to parts missing from the scan.
Within expanded object bounds, $R_i^{\rm mask}$ selects Gaussian
centers whose projections fall inside instance masks in multiple views.
We remove the union over all target objects.

\textbf{Inpaint and complete the exposed background.}
Image inpainting supplies background appearance targets, not 3D geometry.
Each object uses one primary edited view to avoid conflicting independent
inpaintings. We composite all edits sharing a frame into one target,
so one object's supervision does not restore another's original appearance.
Around the exposed region, we fit a robust support plane and initialize
normal-aligned Gaussian disks, colored from the edited view where visible
and neighbouring observations elsewhere.
Refinement against the edited-image targets optimizes only these new
Gaussians, keeping retained source Gaussians fixed.
This produces a shared renderable background under a local planar
assumption, suitable for regions such as tabletops, rather than a unique
recovery of hidden surfaces. Unsupported regions and failed completions
are recorded.

\subsection{Interactive Environment}
\label{sec:assembly}
Registered assets and Background Gaussians form the interactive scene.

\subsubsection{Couple visual and physical state}
Collision meshes determine contacts, the simulator computes body poses,
and Gaussians provide the corresponding visual appearance.
Where supported, CoACD converts completed meshes into convex collision
components~\cite{coacd}; physical parameters come from priors unless
separately measured.

As illustrated in Fig.~\ref{fig:system}(c), visual and collision assets
share an initial metric placement.
For simulator body pose $T_i(t)\in\mathrm{SE}(3)$ and initial visual
placement $S_i(0)$, we update
\begin{equation}
S_i(t)=T_i(t)T_i(0)^{-1}S_i(0).
\label{eq:motion}
\end{equation}
This applies the body's relative motion while preserving the initial
body-to-asset offset, without requiring the visual origin to coincide
with the center of mass. Shared motion does not guarantee identical
visual and collision surfaces.

For scale $s$, rotation $R$, and translation $\mathbf t$, Gaussian means
and covariances transform as
\begin{equation}
\boldsymbol\mu'=sR\boldsymbol\mu+\mathbf t,\qquad
\Sigma'=s^2R\Sigma R^\top.
\label{eq:gaussians}
\end{equation}
Scale is applied once at construction, after which body motion is rigid.
Cameras, bodies, and renderings use the same simulator state.
View-dependent appearance requires consistent viewing directions.
Here, \emph{simulation-ready} means loadable rigid objects with
state-linked appearance, not a guarantee of physical accuracy or task utility.

\subsubsection{Optional appearance harmonization}
\label{sec:harmonization}
Object motion does not update illumination encoded in Gaussian appearance.
After rendering, optional pretrained DiffusionHarmonizer
(Fig.~\ref{fig:teaser}(c))~\cite{harmonizer} takes the current image and
a causal history of enhanced frames from the same camera and episode.
It modifies image appearance without changing Gaussian parameters,
collision geometry, or dynamics, and does not reconstruct a lighting
model or guarantee physically correct shadows.
We retain raw renderings to separate scene-construction quality from
image enhancement.

\section{Experiments}
\label{sec:eval}
\label{sec:results}
\begin{table*}[t]
\centering\footnotesize
\setlength{\tabcolsep}{3.1pt}
\renewcommand{\arraystretch}{1.13}
\caption{\textbf{Coupled Scene Construction on 50 ScanNet++ scenes.} Rows vary its candidate pool, selection, retry, and acceptance rule. Retained counts are out of 1,871 requests. T: TRELLIS; R: ReconViaGen.}
\vskip-2ex
\label{tab:real-scan}
\begin{tabular}{@{}llcccrrrrr@{}}
\toprule
Configuration & Pool & Selection & Retry & Strict & Retained & Matched & F1@20 $\uparrow$ & CD (cm) $\downarrow$ & min/scene \\
\midrule
TRELLIS only & T & Single & No & No & 1640 & 517 & 0.301 & 11.767 & 13.58 \\
Fixed priority & T+R & R first & No & No & 1800 & 566 & 0.336 & 7.638 & 35.68 \\
\name\ without retry & T+R & Evidence & No & No & 1800 & 566 & 0.348 & 7.196 & 35.68 \\
\textbf{\name\ (ours)} & T+R & Evidence & Yes & No & 1800 & 566 & 0.383 & 5.720 & 70.82 \\
\name\ + strict acceptance & T+R & Evidence & Yes & Yes & 399 & 134 & 0.425 & 4.257 & 70.82 \\
\bottomrule
\end{tabular}
\end{table*}

\begin{table*}[t]
\centering\footnotesize
\setlength{\tabcolsep}{3.6pt}
\renewcommand{\arraystretch}{1.13}
\caption{\textbf{Registered-asset execution in the Interactive Environment.}  $S/P$ and $S/E$ denote overall and conditional success.}
\vspace{-1em}
\label{tab:mechanism}
\begin{tabular}{@{}lrrrrrrr@{}}
\toprule
Configuration & Executed & Successful & $S/P$ (\%) & $S/E$ (\%) & To cabinet & To sink & From sink \\
\midrule
Original environment (reference) & 480 & 385 & 80.2 & 80.2 & 119/160 & 130/160 & 136/160 \\
\midrule
TRELLIS only & 310 & 73 & 15.2 & 23.5 & 42/160 & 19/160 & 12/160 \\
Fixed priority & 320 & 131 & 27.3 & 40.9 & 63/160 & 39/160 & 29/160 \\
\name\ without retry & 320 & 118 & 24.6 & 36.9 & 63/160 & 35/160 & 20/160 \\
\textbf{\name\ (ours)} & 320 & 128 & 26.7 & 40.0 & 65/160 & 38/160 & 25/160 \\
\bottomrule
\end{tabular}
\vspace{-2em}
\end{table*}

\subsection{Experimental Setting}
\label{sec:implementation}
\textbf{Implementation.}
SAM3 processes every twelfth training frame using a fixed household-object
vocabulary and a mask-score threshold of 0.45. Observations are merged
on a 2\,cm voxel grid at an overlap threshold of 0.25, requiring two
supporting views and category-dependent extent checks. Registration uses
20K mesh points, $10^\circ$ yaw steps, and 3\,cm distance clipping.
Retry tests five alternative signed source-up axes. Candidates are ranked
lexicographically by check satisfaction, failed-check count, registration
residual, scale discrepancy, and settling displacement, with deterministic
tie-breaking. Gaussian removal uses 3\,cm observed-surface and 2.4\,cm
asset-surface radii with at least two mask votes. Local fill uses a 5\,mm
grid. MuJoCo runs robot rollouts, while PyBullet supports construction
probes and dynamics tests~\cite{mujoco,pybullet}. Each study retains its
shared room collisions or per-object support shims. All experiments are conducted on NVIDIA A6000.

\textbf{Scene evaluation.}
The main study covers 1,871 object requests from 50 ScanNet++
scenes~\cite{scannetpp}. Construction uses training observations, with
reference instances associated after candidate selection. Retention
measures candidate availability over all requests. Geometry uses F1 at
20\,mm and symmetric Chamfer distance (CD, centimeters) on independent
matches without evaluator-side alignment. Export validity and isolated-body
drop stability are evaluated separately from the construction probe used
for selection. Visual fidelity uses PSNR, SSIM, and AlexNet LPIPS on raw
Gaussian renderings~\cite{ssim,lpips}. Construction time includes generation,
registration, checks, and retries, but excludes acquisition,
source-Gaussian training, and queueing.

\textbf{Interaction evaluation.}
RoboCasa provides 48 configurations across eight layouts and three task
families, with ten paired resets per configuration, giving 480 planned
trials per method~\cite{robocasa365}. Ideal posed RGB-D observations
reconstruct the target, while the room and destination remain native.
Policy observations use uniform-material native mesh rendering without
Harmonizer. All methods share the robot, controller, instructions, cameras,
horizons, and native success tests. Frozen OpenPI
\texttt{pi05\_pretrain\_human300} uses three $224\times224$ camera inputs
and five-step replanning. Fresh policy processes use matched sampling
streams and saved robot, object, controller, and sensor states.
Perturbations affect estimated placements, not reference poses.
Unchanged-asset import controls check states, observations, and task
predicates before replacement. The scene set was previously evaluated,
rather than newly held out.

For $N_p$ planned trials, $N_e$ executions, and $N_s$ successes,
\begin{equation}
\underbrace{\frac{N_s}{N_p}}_{\text{overall success}}=
\underbrace{\frac{N_e}{N_p}}_{\text{coverage}}\,
\underbrace{\frac{N_s}{N_e}}_{\text{conditional success}}.
\label{eq:coverage}
\end{equation}
Construction rejections contribute no successes to fully accounted
studies, while missing evaluations remain unmeasured. Policy intervals
use paired hierarchical resampling over layouts, configurations, and
resets, rather than treating resets as independent reconstructions.

\subsection{Quantitative Results}
\label{sec:quantitative}

\textbf{Scene construction and physical validity.}
Table~\ref{tab:real-scan} compares TRELLIS-only generation with a
combined TRELLIS/ReconViaGen pool, varying selection, registration
retry, and strict acceptance. Fixed priority selects ReconViaGen
when available and TRELLIS otherwise. With the same 1,800 retained
candidates and 566 independent matches, evidence-based selection
increases F1 from 0.336 to 0.348, and retry raises it to 0.383.
The respective paired gains have 95\% intervals of [0.001, 0.022]
and [0.007, 0.075]. Relative to fixed priority, the full constructor
reduces CD by 25.1\%, although retry increases construction time
from 35.68 to 70.82 minutes per scene. These gains reflect improved
candidate choice and alignment at fixed retention, whereas the
TRELLIS-only comparison also changes the candidate pool and retention. Strict acceptance rejects candidates that fail construction checks,
yielding higher conditional F1 on a smaller matched population.
However, 17/134 accepted matches still exhibit geometry collapse,
and 361 of its 394 verified exports pass the independent drop test.

\begin{figure*}[htbp]
    \centering
    \includegraphics[width=0.95\linewidth]{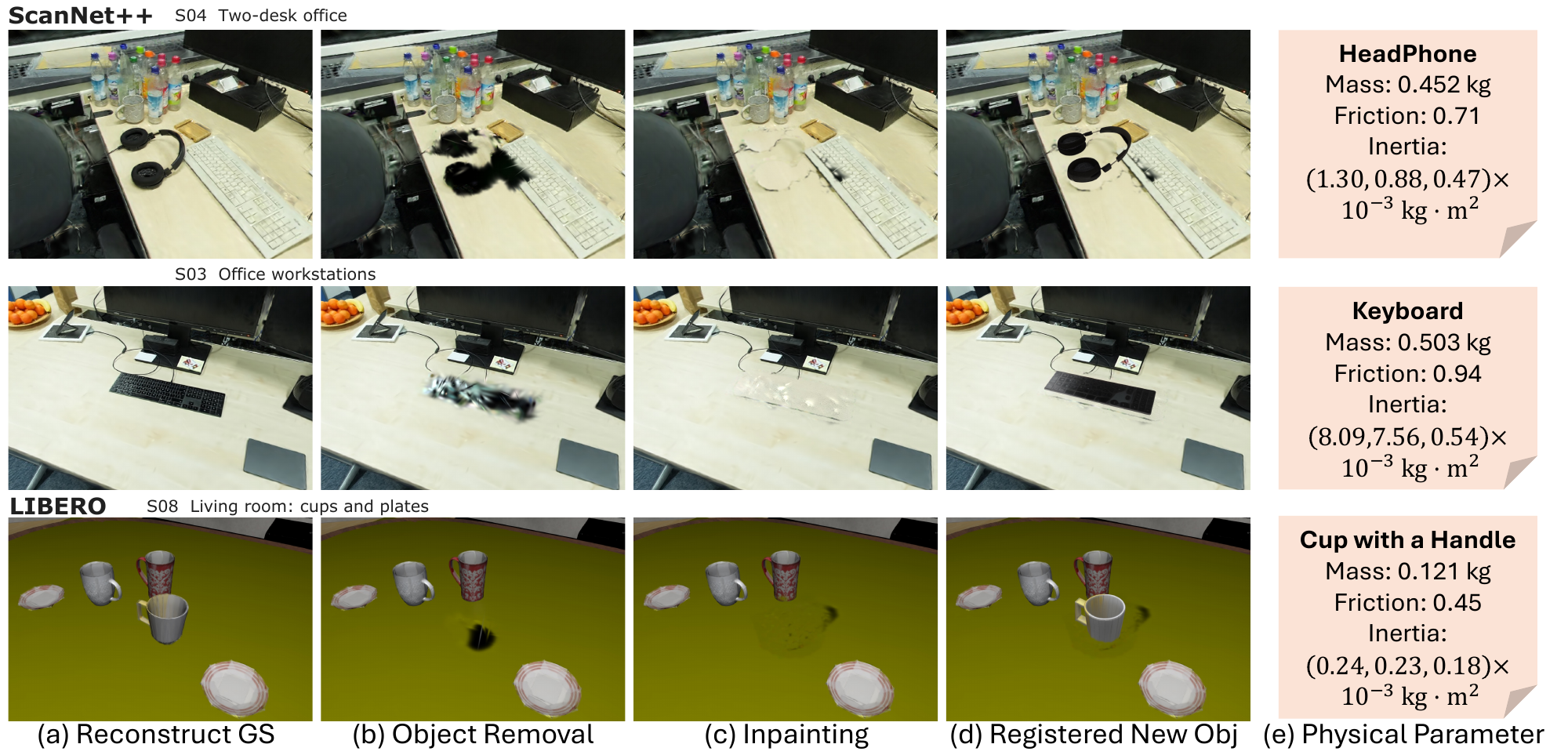}
    \vspace{-1em}
    \caption{\textbf{Qualitative scene conversion with \name.}
    Examples from ScanNet++ (top two rows) and LIBERO (bottom row) show
    (a) the source Gaussian reconstruction, (b) object removal,
    (c) background inpainting, (d) the registered replacement asset,
    and (e) its assigned mass, friction, and inertia.}
    \label{fig:qualitative-results}

\end{figure*}

\begin{table}[t]
    \centering
    \footnotesize
    \setlength{\tabcolsep}{4pt}
    \renewcommand{\arraystretch}{1.1}
    \caption{\textbf{Visual fidelity on ScanNet++.}}
    \vspace{-1em}
    \label{tab:scan-appearance}
    \begin{tabular}{@{}lrrr@{}}
        \toprule
        Representation & PSNR $\uparrow$ & SSIM $\uparrow$
        & LPIPS $\downarrow$ \\
        \midrule
        Source Gaussians
        & 21.582 & 0.8444 & 0.2993 \\
        \name{} (factorized)
        & 20.438 & 0.8318 & 0.3168 \\
        \bottomrule
    \end{tabular}
    \vspace{-2em}
\end{table}

\textbf{Visual fidelity.}
Conversion incurs a 1.144\,dB PSNR loss
(Table~\ref{tab:scan-appearance}), quantifying the joint visual cost
of object replacement and background completion in the initial state.
Surfaces exposed only after motion are not evaluated here.

\textbf{Controlled manipulation.}
In Table~\ref{tab:mechanism}, \name{} succeeds on 128/480 planned trials,
compared with 73/480 for TRELLIS only. The paired gain is 11.46 percentage
points, with a 95\% interval of [4.38, 19.79], and improvements occur in
all three task families. Fixed priority, however, achieves 131 successes,
and the intervals for adding selection and then retry both cross zero.
Thus, the expanded constructor improves manipulation relative to the
single-generator configuration, but the geometric gains from selection
and retry do not establish additional policy gains. The geometry and
policy studies also use different objects.

The original environment succeeds on 385/480 trials. For \name{},
the 352 unsuccessful planned trials comprise 160 without an executable
construction and 192 failures after execution. Both construction
availability and interaction quality limit performance. This experiment
evaluates registered-asset replacement under native mesh observations,
separately from the Gaussian-rendered demonstrations below.

\subsection{Qualitative Results}
\label{sec:qualitative}

\textbf{Coupled object and background construction.}
Figure~\ref{fig:qualitative-results} follows the conversion from a captured
Gaussian scene through source removal and background inpainting to a
registered asset with physical parameters. The sequence illustrates that the asset being inserted corresponds to
the source appearance being removed, while completion supplies the
background exposed by that removal. The parameter panel reports assigned
physical properties. 
\begin{figure}[t]
    \centering
    \includegraphics[width=0.95\linewidth]{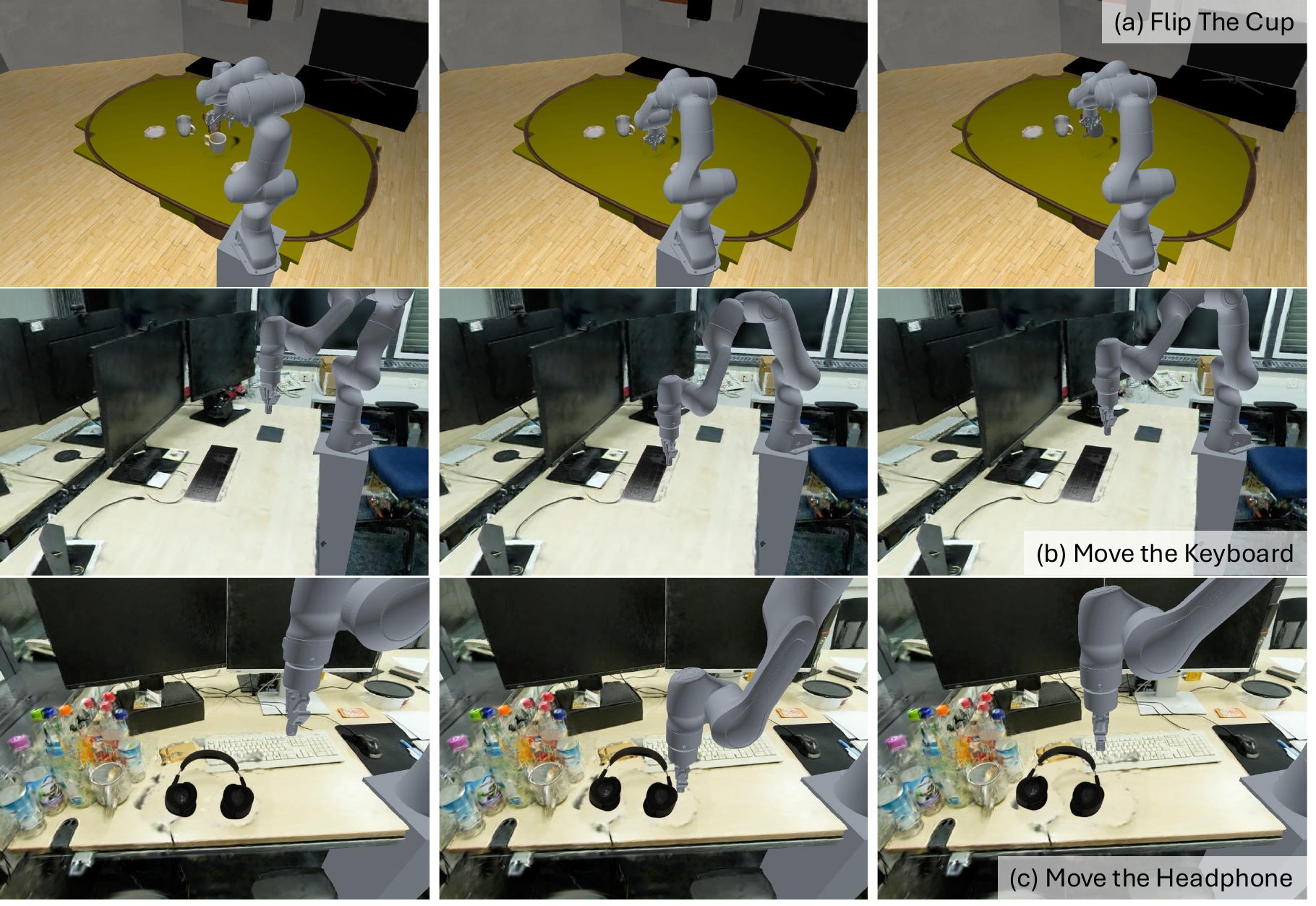}
    \caption{\textbf{Robot manipulation in the Interactive Environment.}
    Sequences show (a) flipping a cup, (b) moving a keyboard, and
    (c) moving headphones. Frames progress from left to right.}
    \vspace{-1em}
    \label{fig:qualitative-robot}
\end{figure}

\begin{figure}[t]
    \centering
    \includegraphics[width=0.95\linewidth]{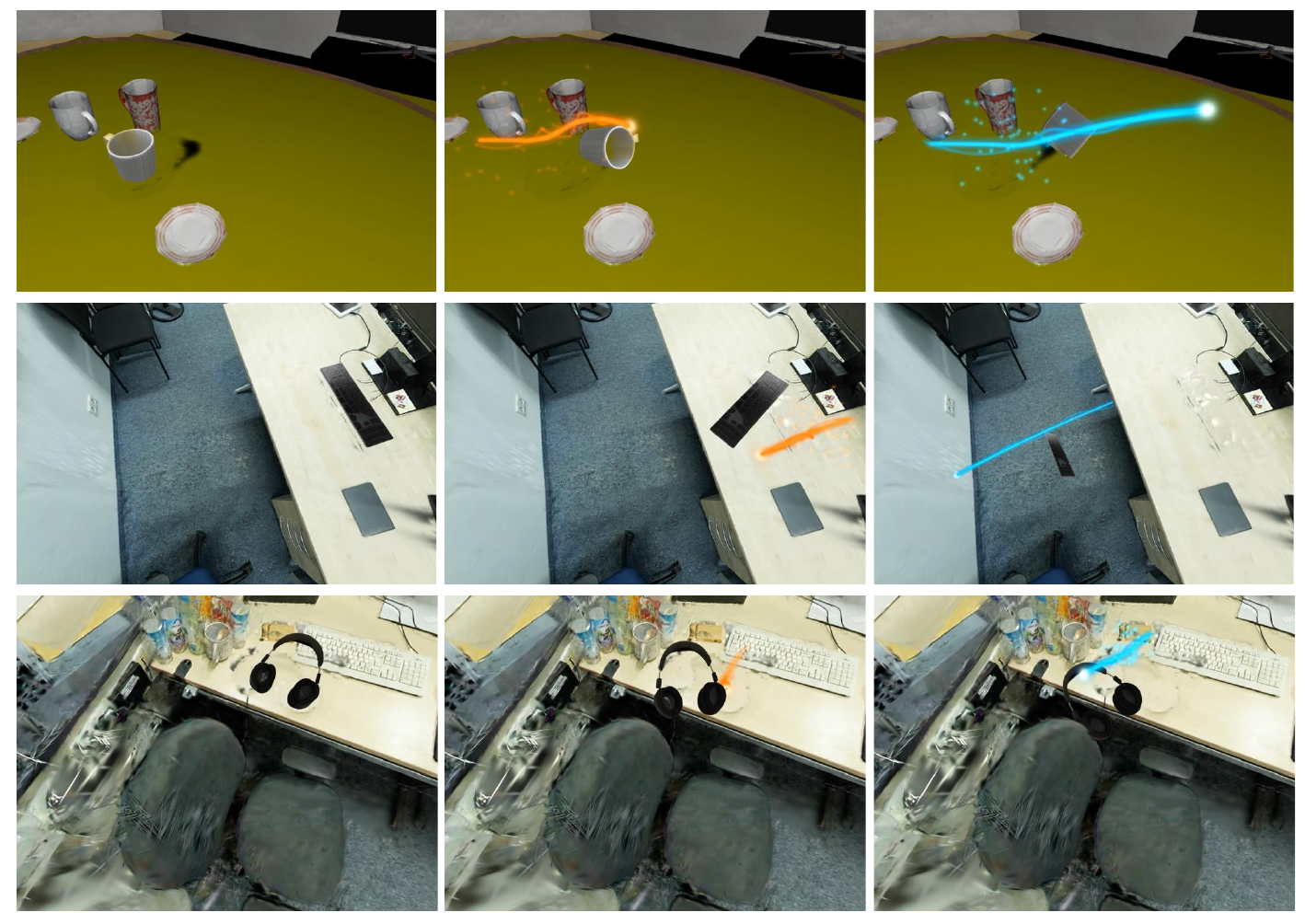}
    \caption{\textbf{Interactive object perturbations in \name.}
    From top to bottom, shooting interactions displace and rotate
    a cup, a keyboard, and headphones.
    Each row shows the initial configuration followed by two interaction
    states, with colored trails indicating the shots.}
    \label{fig:qualitative-shooting}
\end{figure}

\textbf{Manipulation and object motion.}
Figure~\ref{fig:qualitative-robot} illustrates cup flipping and the movement
of a keyboard and headphones. Figure~\ref{fig:qualitative-shooting}
complements these robot interactions with object displacement and rotation
under shooting interactions. Together, they illustrate independently
movable content in the Interactive Environment and the simulator-linked
rendering described in Sec.~\ref{sec:assembly}. These demonstrations
are separate from the paired RoboCasa policy evaluation in
Table~\ref{tab:mechanism} and do not measure visual--collision pose error.

\begin{figure}[t]
    \centering
    \includegraphics[width=0.95\linewidth]{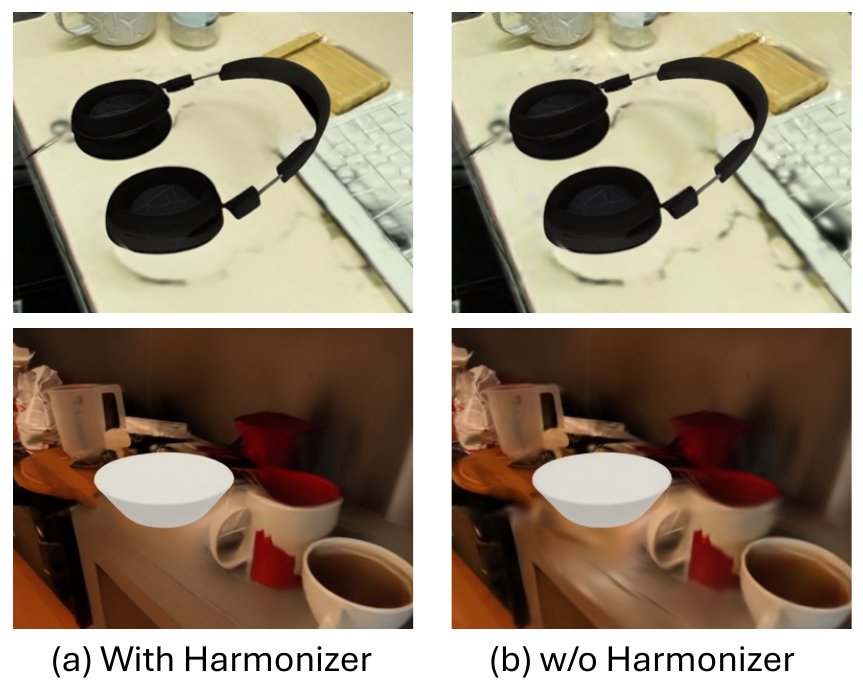}
    \caption{\textbf{Optional appearance harmonization.}
    Renderings with and without the post-rendering Harmonizer.}
    \label{fig:quali_harmonizer}
\end{figure}

\textbf{Appearance harmonization.}
Figure~\ref{fig:quali_harmonizer} compares rendered observations with and
without the optional Harmonizer in Sec.~\ref{sec:harmonization}.
Unlike background completion, which constructs scene Gaussians, this
stage modifies rendered images without changing geometry or simulator
state. It addresses the appearance mismatch discussed in
Fig.~\ref{fig:teaser}(c). 

\subsection{Ablations}
\label{sec:ablations}
\textbf{Object generation backend.}
\begin{table}[t]
\centering\footnotesize
\setlength{\tabcolsep}{2.0pt}
\renewcommand{\arraystretch}{1.18}
\caption{\textbf{Object-candidate generators within \name.}}
\label{tab:generator-backends}
\begin{tabular}{@{}lrrrrr@{}}
\toprule
Generator & Exported & F1@20 $\uparrow$ & F1@40 $\uparrow$ & CD (cm) $\downarrow$ & Stable \\
\midrule
TRELLIS & 29/32 & \textbf{0.3087} & \textbf{0.5311} & \textbf{7.1939} & 15/29 \\
TRELLIS.2 & 29/32 & 0.2305 & 0.4482 & 10.2237 & 10/29 \\
SAM 3D Objects & 29/32 & 0.1847 & 0.3724 & 9.4079 & \textbf{16/29} \\
ReconViaGen & 29/32 & 0.2883 & 0.5064 & 8.6694 & 14/29 \\
\bottomrule
\end{tabular}
\end{table}

We compare TRELLIS, TRELLIS.2, SAM 3D Objects, and
ReconViaGen~\cite{trellis,trellis2,sam3dobjects,reconviagen} on
32 ScanNet++ development objects, with inputs available for 29.
Single-image methods share a frame and mask, while ReconViaGen uses
5--12 training views. All candidates share observation-only
signed-source-up registration, mesh simplification, CoACD decomposition,
and URDF import without quality-based rejection. Geometry uses ten
common independent matches, reporting F1 at 20/40\,mm and CD without
evaluator-side alignment.

All 29 exports per backend undergo a 240\,Hz plane-drop test with
two seconds of velocity-zeroed settling followed by two seconds of
free dynamics. Stability requires link-frame drift below 3\,cm and
final link height at least $-5$\,cm. Collision meshes have a
40K-triangle budget, with mass 0.3\,kg, friction 0.5, and restitution 0.0. All backends pass collision import (Table~\ref{tab:generator-backends}).
TRELLIS achieves the highest F1 and lowest CD, whereas SAM 3D Objects
has the highest observed settling count (16/29). No backend leads both
geometry and stability, which are evaluated on different populations.
This single-seed study compares individual backends, not selection
over a four-generator pool.

\textbf{Background inpainting backend.}
\begin{table}[t]
\centering\footnotesize
\setlength{\tabcolsep}{2.5pt}
\renewcommand{\arraystretch}{1.13}
\caption{\textbf{Inpainting targets for Background Gaussians.}}
\label{tab:background}
\begin{tabular}{@{}lrrrrr@{}}
\toprule
Editor & Meas. & PSNR & SSIM & LPIPS & Outside MAE \\
\midrule
Telea~\cite{telea2004inpainting} & 33 & 26.45 & 0.771 & 0.162 & 0.000 \\
SDXL~\cite{podell2024sdxl} & 33 & 22.27 & 0.712 & 0.174 & 0.016 \\
\shortstack[l]{Gemini 3.1\\Flash Image} & 32 & 26.81 & 0.749 & 0.216 & 0.013 \\
\bottomrule
\end{tabular}
\end{table}

We compare Telea, SDXL, and Gemini using identical images, masks,
and compositing, with clean same-state targets reserved for evaluation.
Of 48 planned cases across 12 instances, all metrics use the same
32 cases from nine instances, averaged first within and then across
instances. In Table~\ref{tab:background}, \emph{Meas.} counts available
outputs. PSNR evaluates the masked hole, SSIM/LPIPS evaluate the crop,
and Outside MAE measures changes outside the mask before compositing. Gemini achieves the highest hole PSNR, exceeding Telea by 0.36\,dB,
whereas Telea gives higher crop SSIM, lower LPIPS, and zero outside-mask
error. Editor rankings therefore depend on the evaluation region and
metric. These results assess 2D completion targets, not the reconstructed
3D background.

\textbf{Input requirements.}
\begin{table}[t]
\centering
\footnotesize
\setlength{\tabcolsep}{2.7pt}
\renewcommand{\arraystretch}{1.13}
\caption{\textbf{Input requirements on ScanNet++.} Separate tiered study varies discovery and surface input. }
\label{tab:input-requirements}
\begin{tabular}{@{}lrrr@{}}
\toprule
Instances / surface input & Count & Yield & F1@20 \\
\midrule
Annotated / scan mesh & 789/50 & 58.0\% & 0.708 \\
Automatic / scan mesh & 1,082/50 & 62.0\% & 0.582 \\
Annotated / splat-derived & 678/49 & 56.0\% & 0.693 \\
Automatic / splat-derived & 930/49 & 59.0\% & 0.553 \\
Single-image configuration & 389/40 & 17.2\% & 0.309 \\
\bottomrule
\end{tabular}

\end{table}

We vary instance annotations, scene-surface input, and multi-view
availability, using single-view TRELLIS generation throughout.
Splat-derived surfaces fuse rendered training depths with 5\,mm TSDF
voxels. This separate study accepts candidates using a size check and
construction-surface F1 thresholds of 0.40 at 20\,mm or 0.20 at 40\,mm.
Annotated rows use reference instance surfaces during registration.
Automatic instances are scored against independently matched references.

With annotated instances, replacing the scan mesh with a splat-derived
surface reduces F1 by 0.015 and yield by two percentage points
(Table~\ref{tab:input-requirements}). With automatic discovery, matched
F1 changes from 0.582 to 0.553. These results support using Gaussian-derived
geometry, although the evaluated populations differ. Against their own
construction surfaces, the automatic variants score 0.630 and 0.667,
respectively, reversing the independently evaluated ranking.
Construction agreement is therefore not a substitute for reference
accuracy. The single-image configuration retains 17.2\% of requests
with conditional F1 of 0.309, but changes masks, image coverage, and
observed geometry together.

\subsection{Discussions}
\label{sec:limits}
\label{sec:exp-discussions}
\textbf{Component evidence.}
Experiments test selection, registration retry, and backend choices,
while the input study measures construction quality, not discovery
precision or recall. Source-Gaussian removal, local 3D fill, and
simulator-linked appearance have qualitative evidence
(Figs.~\ref{fig:qualitative-results}--\ref{fig:qualitative-shooting}),
but no isolated quantitative tests. 

\textbf{Interaction scope.}
In a separate 48-configuration study with shared repair actions,
adaptive and fixed-order contextual repair both yield 65/480 successes
versus 121/480 without repair. Conditional success rises from 37.8\%
to 54.2\%, but executions fall from 320 to 120. Separately, destination
reconstruction reduces successes from 45/240 to 20/240 across
24 configurations and four layouts, retaining native articulation
and goals. These results expose coverage and asset-compatibility
limitations. Contextual repair differs from alignment-only retry,
and policy success does not establish real-world dynamic fidelity.

\textbf{Capture and physical scope.}
The exploratory TRELLIS-only RGB-video study yields seven loadable
targets from 24 instances and 21/70 successes versus 53/70 for the
reference, with 170 method trials unmeasured. Association, calibration,
and appearance issues, plus a post-failure marker-estimator revision,
limit interpretation. The pipeline assumes rigid objects, prior
physical parameters, and locally planar fill. Articulation, deformables,
identified dynamics, arbitrary hidden geometry, and physical-robot
transfer remain outside the demonstrated scope.

\section{Conclusion}
\label{sec:conclusion}
We presented \name, a Gaussian-native pipeline that converts selected
objects into movable simulator assets while preserving unedited scene
content. Shared object evidence couples asset construction, source removal,
and background completion, while simulator poses drive Gaussian appearance.
Experiments show improved geometry at fixed retention and manipulation
gains over a single-generator baseline in controlled asset-replacement
tests, alongside a visual-fidelity trade-off. This coupling extends
captured Gaussian scenes beyond static rendering to object-level
interaction. 

\noindent\textbf{Generative AI use disclosure.}
OpenAI Codex was used to assist with drafting and debugging portions of the experimental code and with language editing throughout the manuscript. The authors reviewed and validated the resulting code and text. All reported results were obtained from actual experiment runs and verified by the authors; no empirical result values were invented, altered, or synthesized by AI.

\bibliographystyle{IEEEtran}
\bibliography{refs,refs_pipeline}
\end{document}